\documentclass{article}

\usepackage{arxiv}

\usepackage[utf8]{inputenc} 
\usepackage[T1]{fontenc}    
\usepackage{hyperref}       
\usepackage{url}            
\usepackage{booktabs}       
\usepackage{amsfonts}       
\usepackage{nicefrac}       
\usepackage{microtype}      
\usepackage{lipsum}		
\usepackage{graphicx}
\usepackage{natbib}
\usepackage{doi}
\usepackage{amsmath}
\usepackage{amssymb}
\usepackage{enumitem}

\title{On equivalence between linear-chain conditional random fields and hidden Markov chains}

\author{
\href{https://orcid.org/0000-0003-3595-0826}
{\includegraphics[scale=0.06]{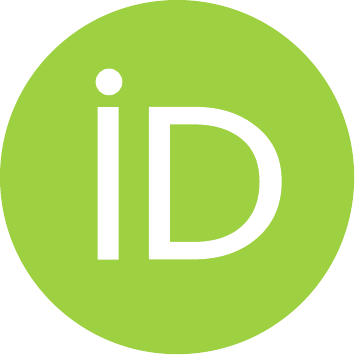}\hspace{1mm}
Elie~Azeraf}
\thanks{Elie Azeraf is also a member of Telecom SudParis, Institut Polytechnique de Paris.} \\
Watson Department \\
IBM France \\
Paris, France \\
\texttt{elie.azeraf@ibm.com} \\
\And
\href{https://orcid.org/0000-0002-7648-2515}
{\includegraphics[scale=0.06]{orcid.pdf}\hspace{1mm}
Emmanuel~Monfrini} \\
Telecom SudParis \\
Institut Polytechnique de Paris \\
Paris, France \\
\texttt{emmanul.monfrini@telecom-sudparis.eu} \\
\And
\href{https://orcid.org/0000-0002-1371-2627}
{\includegraphics[scale=0.06]{orcid.pdf}\hspace{1mm}
Wojciech~Pieczynski} \\
Telecom SudParis \\
Institut Polytechnique de Paris \\
Paris, France \\
\texttt{wojciech.pieczynski@telecom-sudparis.eu} \\
}

\date{}

\begin{document}
\maketitle

\begin{abstract}
Practitioners successfully use hidden Markov chains (HMCs) in different problems for about sixty years. HMCs belong to the family of generative models and they are often compared to discriminative models, like conditional random fields (CRFs). Authors usually consider CRFs as quite different from HMCs, and CRFs are often presented as interesting alternative to HMCs. In some areas, like natural language processing (NLP), discriminative models have completely supplanted generative models. However, some recent results show that both families of models are not so different, and both of them can lead to identical processing power. In this paper we compare the simple linear-chain CRFs to the basic HMCs. We show that HMCs are identical to CRFs in that for each CRF we explicitly construct an HMC having the same posterior distribution. Therefore, HMCs and linear-chain CRFs are not different but just differently parametrized models.
\end{abstract}


\section{Introduction}

Let $Z_{1:N} = (Z_1, ..., Z_N)$ be a stochastic sequence, with $Z_n = (X_n, Y_n)$. $X_1, ..., X_N$ take their values in a finite set $\Omega$, while $Y_1, ..., Y_N$ take their values in a
discrete set $\Lambda$. Realizations of $X_{1:N}=(X_1, ..., X_N)$ are hidden while realizations of $Y_{1:N}=(Y_1, ..., Y_N)$ are observed, and the problem we deal with is to estimate $X_{1:N} = x_{1:N}$ from $Y_{1:N} = y_{1:N}$.

The simplest model allowing dealing with the problem is the well-known hidden Markov chain (HMC). In spite of their simplicity, HMCs are very robust and provide quite satisfactory results in many applications. We only cite the pioneering papers (\cite{baum1970maximization,rabiner1989tutorial}), and some books (\cite{10.5555/1088883,koski2001hidden}), among great deal of publications. However, they can turn out to be too simple in complex cases and thus authors extended them in numerous directions. In particular, conditional random fields (\cite{lafferty2001conditional,sutton2006introduction}) are considered as interesting alternative to HMCs, especially in Natural Language Processing (NLP) area. Used in different areas as diagnostic (\cite{fang2018novel,fang2019real}), natural language processing (\cite{jurafsky2000speech}), entity recognition (\cite{song2019named}), or still relational learning (\cite{sutton2006introduction}). In general, authors consider CRFs as quite different from HMCs, and often prefers the former to the latter. In this paper, we show that CRFs and HMCs may be not so different. More precisely, we show that basic linear- chain CRFs are equivalent to HMCs.

Let us specify what “equivalence” in the paper’s title means. One can notice that HMCs and CRFs cannot be compared directly as they are of different nature. Assuming that a “model” is a distribution $p(x_{1:N}, y_{1:N})$, we may say that HMC is a model, while CRF is a family of models, in which all models have the same $p(x_{1:N} | y_{1:N})$, but can have any $p(y_{1:N})$. We will say that a CRF $p(x_{1:N} | y_{1:N})$ is equivalent to a HMC $q(x_{1:N}, y_{1:N})$ if and only if $p(x_{1:N} | y_{1:N}) = q(x_{1:N} | y_{1:N})$. To show that linear-chain CRFs are equivalent to HMCs it is thus sufficient to show that for each linear-chain CRF $p(x_{1:N} | y_{1:N})$, it is possible to find a HMC $q(x_{1:N}, y_{1:N})$ such that $p(x_{1:N} | y_{1:N}) = q(x_{1:N} | y_{1:N})$. This precisely is the contribution of the paper.

More generally, let us note that certain criticisms of the HMCs, put forward to justify the preference of the CRFs, currently appear to be not always entirely justified. For example, in monitoring problems, two independence conditions inherent to HMCs were put forward to justify this preference. However, these conditions are sufficient conditions for Bayesian processing, not necessary ones. Indeed, it is possible removing those considering Pairwise Markov Chains (PMCs), which extend HMCs and allow the same Bayesian processing (\cite{pieczynski2003pairwise,gorynin2018assessing}). Another example is related to NLP. HMCs are considered as generative models, and as such improper to NLP because of the fact $p(x_{1:N} | y_{1:N})$ are difficult to handle to (\cite{jurafsky2000speech, brants2000tnt, mccallum2000maximum}). However, as recently shown in \cite{azeraf2020hidden}, while defining Bayesian processing methods HMCs can also be used in discriminative way, without calling on $p(y_{1:N} | x_{1:N})$. The same is true in the case of other generative models like Naïve Bayes \cite{azeraf2021highly}.

\section{Linear-chain CRF and HMC}

\subsection{Bayesian classifiers}

In the Bayesian framework we consider, there is a loss function  $L(x_{1:N}^\ast,x_{1:N})$, where $x_{1:N}$ is the true value and $x_{1:N}^\ast$ is the estimated one. Bayesian classifier $y_{1:N} \rightarrow{{\hat{x}}_{1:N}=\hat{s}}_B^L(y_{1:N})$ is optimal in that it minimizes the mean loss $\mathbb{E}\left[ L(\hat{s} \left( Y \right), X) \right]$. It is defined with
\begin{align}
{{\hat{x}}_{1:N} = \hat{s}}_B^L \left( y_{1:N} \right) = 	
\arg \inf_{x_{1:N}^*} \mathbb{E}[ L( x_{1:N}^*, X_{1:N})| Y_{1:N} = y_{1:N}], 
\end{align}
where $\mathbb{E}[L(x_{1:N}^\ast,X_{1:N})|Y_{1:N}]$ denotes the conditional expectation. In this paper, we consider the Bayesian classifier ${\hat{s}}_B^L$ corresponding to the loss function 
\begin{align}
    L\left(x_{1:N}^\ast,x_{1:N}\right) = 1(x_1^* \neq x_1) + ... + 1(x_N^* \neq x_N),
\end{align}
which simply means that the loss is the number of wrongly classified data. Called « maximum posterior mode » (MPM), the related Bayesian classifier is defined with
\begin{align}
    \left[{\hat{x}}_{1:N} = \left({\hat{x}}_1, ..., {\hat{x}}_N \right) = {\hat{s}}_B^L \left( y_{1:N} \right) \right] \Longleftrightarrow [\forall n=1, ..., N, p({{\hat{x}}_n\left|y_{1:N}\right)} = \sup_{x_n}( p\left( x_n | y_{1:N} \right))] 
\end{align}

Let us remark that Bayesian classifiers ${\hat{s}}_B^L$ only depends on $p(x_{1:N}|y_{1:N})$, and are independent from $p(y_{1:N})$. In other words, for any distribution $q(y_{1:N})$, every other law of $(X_{1:N},\ Y_{1:N})$ of the form $q\left(x_{1:N} | y_{1:N} \right) = p(x_{1:N} | y_{1:N}) q(y_{1:N})$ gives the same Bayesian classifier ${\hat{s}}_B^L$. This shows that dividing classifiers into two categories “generative” and “discriminative” as usually done is somewhat misleading as they all are discriminative. Such a distinction is thus related to the way classifiers are defined, not to their intrinsic structure.  

\subsection{Equivalence between linear-chain CRF and a family of HMCs}

We show in this section that for each linear-chain CRF one can find an equivalent HMC, with parameters specified from the considered CRF.

The following general Lemma will be useful in the sequence:

\subsubsection*{Lemma} 

Let $W_{1:N} = (W_1, ..., W_N)$ be random sequence, taking its values in a finite set $\Omega$. Then
\begin{enumerate}[label=(\roman*)]
    \item $W_{1:N}$ is Markov chain iff there exist $N - 1$ functions $\varphi_1,\ \ldots,\ \varphi_{N-1}$ from $\Omega^2$ to $\mathbb{R}^+$ such that
    \begin{align}
        p(w_1, ..., w_N) \propto \varphi_1(w_1, w_2) ... \varphi_{N - 1}(w_{N - 1}, w_N),
        \label{eq_4}
    \end{align}
    where “$\propto$” means “proportional to”;
    
    \item  for HMC defined with $\varphi_1, ..., \varphi_{N-1}$ verifying (\ref{eq_4}), $p(w_1)$ and $p(w_{n + 1} |w_n)$ are given with
    \begin{align}
    \begin{split}
        p(w_1) &= \frac{\beta_1(w_1)}{\sum_{w_1}{\beta_1(w_1)}}; \\
        p(w_{n + 1} |w_n) &= \frac{\varphi_n (w_n, w_{n+1}) \beta_{n+1}(w_{n+1})}{\beta_n(w_n)} ,
    \end{split}
    \end{align}
    where $\beta_1(w_1), ..., \beta_N(w_N)$ are defined with the following backward recursion:
    \begin{align}
    \begin{split}
        \beta_N(w_N) &= 1, \\
        \beta_n(w_n) &= \sum_{w_{n + 1}} \varphi_n(w_n, w_{n+1}) \beta_{n + 1}(w_{n + 1})
    \end{split}
    \end{align}
    For the proof see (\cite{lanchantin2011unsupervised}), Lemma 2.1, page 6.
\end{enumerate}
We can state the following Proposition.

\subsubsection*{Proposition}

Let $Z_{1:N}=(Z_1, ..., Z_N)$ be stochastic sequence, with $Z_n=(X_n, Y_n)$. Each $(X_n, Y_n)$ takes its values in $\Omega \times \Lambda$, with $\Omega$ and $\Lambda$ finite. If $Z_{1:N}$ is a linear-chain conditional random field (CRF) with the distribution $p(x_{1:N} | y_{1:N})$ defined with
\begin{align}
    p(x_{1:N} | y_{1:N}) = \frac{1}{\kappa(y_{1:N})} exp\left[ \sum_{n = 1}^{N - 1}{V_n(x_n, x_{n + 1})} + \sum_{n = 1}^{N}{U_n(x_n, y_n)} \right],
    \label{eq_7}
\end{align}
where $U_n$ and $V_n$ are arbitrary “potential functions”. Then (\ref{eq_7}) is the posterior distribution of the HMC
\begin{align}
    q(x_{1:N}, y_{1:N}) = q_1(x_1) q_1(y_1 | x_1) \prod_{n = 2}^{N} q_n(x_n |x_{n - 1}) q_n(y_n | x_n), 
    \label{eq_8}
\end{align}
defined as follows.

Let
\begin{align}
    \psi_n(x_n) &= \sum_{y_n} \exp(U(x_n, y_n)) \label{eq_9} \\
    \varphi_1(x_1, x_2) &= \exp(V_1(x_1, x_2)) \psi_1(x_1) \psi_2(x_2);
\end{align}
and, for $n = 2, ..., N-1$: 
\begin{align}
    \varphi_n(x_n, x_{n+1}) = \exp(V_n(x_n, x_{n + 1})) \psi_{n + 1}(x_{n+1}).
\end{align}
Besides, let
\begin{align}
\begin{split}
    \beta_N(x_N) &= 1, \text{ and } \\
    \beta_n(x_n) &= \sum_{x_{n+1}} \varphi_n(x_n, x_{n + 1}) \beta_{n+1}(x_{n+1}) 
\end{split} 
\label{eq_12}
\end{align}
and, for $n = N - 1, ..., 2$.

Then $q(x_{1:N}, y_{1:N})$ is given with
\begin{align}
    q(x_1) &= \frac{\beta_1(x_1)}{\sum\limits_{x_1} \beta_1(x_1)}; \label{eq_13} \\
    q(x_{n + 1} | x_n) &= \frac{\varphi_n(x_n, x_{n+1})\beta_{n + 1}(x_{n + 1}) }{\beta_n(x_n)} \label{eq_14} \\
    q(y_n | x_n) &= \frac{\exp(U(x_n, y_n)}{\psi_n(x_n)}.
    \label{eq_15}
\end{align}

\subsubsection*{Proof}

According to (\ref{eq_9})-(\ref{eq_15}), the distribution (\ref{eq_7}) can be written:
\begin{align*}
    p(x_{1:N} | y_{1:N}) = \frac{n = 1}{\kappa(y_{1:N})} \prod_{n = 1}^{N - 1} \varphi_n(x_n, x_{n + 1}) \prod_{n = 1}^{N} q(y_n | x_n)
\end{align*}
According to the Lemma, $\prod\limits_{n = 1}^{N - 1} \varphi_n(x_n, x_{n+1})$ is a Markov chain defined by (\ref{eq_13}) and (\ref{eq_14}), with $\beta_n(x_n)$ defined (\ref{eq_12}), which ends the proof.

\subsection{HMCs in natural language processing}

Let us notice that relationship between linear-chain CRFs and HMCs have been pointed out and discussed by some authors in the frame of natural language processing (NLP). For example, in (\cite{sutton2006introduction}) authors remark that in linear-chain CRFs it is possible to compute the posterior margins $p(x_n | y_{1:N})$ using the same forward-backward method as in HMCs. However, they keep on saying that CRFs are more general and better suited for applications in NLP. In particular, they consider that CRFs are able to model any kind of features while HMCs cannot. Similarly, in (\cite{jurafsky2000speech}), paragraph 8.5, authors recall that in general it’s hard for generative models like HMCs to add arbitrary features directly into the model in a clean way. 

These arguments are no longer valid since the results presented in (\cite{azeraf2020hidden}). Indeed, according to the results the inability to take into account certain features is not due to the structure of HMCs, but is due to the way of calculating the \textit{a posteriori} laws. More precisely, replacing the classic forward- backward computing by an “entropic” one allows HMCs to take into account the same features as discriminative linear-chain CRFs do. Similar kind of results concerning Naïve Bayes is specified in (\cite{azeraf2021using}).

Let us notice that HMCs defined with (\ref{eq_8}) are even more general than linear-chain CRFs defined with (\ref{eq_7}). Indeed, in the latter we have $p(x_{1:N} | y_{1:N}) > 0$,
while in the former $q(x_{1:N} | y_{1:N}) \geq 0$. However, this is not a very serious advantage as one could extend (\ref{eq_7}) by removing the function $\exp$ and by considering $p(x_{1:N} | y_{1:N}) = \frac{1}{\kappa(y_{1:N})}  \prod\limits_{n = 1}^{N - 1}{V_n(x_n, x_{n + 1})} \prod\limits_{n = 1}^{N}{U_n(x_n, y_n)}$ with all $V_n(x_n, x_{n + 1})$ and $U_n(x_n, y_n)$ positive or null.

\section{Conclusion and perspectives}

We discussed relationships between simple linear- chain CRFs and HMCs. We showed that for each linear-chain CRF, which is a family of models, one can find an HMCs giving the same posterior distribution. In addition, the related HMC’s parameters can be computed from those of CRFs. In particular, joint to results in (\cite{azeraf2020hidden}), this shows that HMCs can be used in NLP with the same efficiency as CRFs do.

Let us mention some perspectives for further work. One recurrent argument in favour of CRFs with respect to HMCs is related to some independence properties assumed in HMCs and considered as binding. More precisely, in HMCs we have $p(y_n | x_{1:N}) = p(y_n | x_n)$ and $p(x_{n + 1} | x_n, y_n) = p(x_{n + 1} | x_n)$. These constraints can be removed by extending HMCs to pairwise Markov chains (PMCs) (\cite{pieczynski2003pairwise, gorynin2018assessing, azeraf2021highly}). More general that HMCs, PMCs allow strictly the same Bayesian processing. Furthermore, PMCs can be extended to triplet Markov chains (TMCs) (\cite{boudaren2014phasic, gorynin2018assessing}), still allowing same Bayesian processing.

Extending HMCs considered in this paper to PMCs and TMCs should lead to extensions of recent hidden neural Markov chain (\cite{icaart21}), which is a first perspective for further works. Of course, there exist many CRFs much more sophisticated that the linear-chain CRF considered in the paper. Let us cite some recent papers (\cite{siddiqi2021improved, song2019named, quattoni2007hidden, kumar2003discriminative, saa2012latent}), among others. Comparing different sophisticated CRFs to different PMCs and TMCs will undoubtedly be an interesting second perspective.

\bibliographystyle{unsrtnat}
\bibliography{references}

\end{document}